\documentclass[10pt,twocolumn,letterpaper]{article}

\usepackage{cvpr}

\usepackage{amsmath}
\usepackage{amssymb}
\usepackage{array}
\usepackage{bbm}
\usepackage{booktabs}
\usepackage{caption}
\usepackage{wrapfig}
\usepackage{epsfig}
\usepackage{graphicx}
\usepackage{multirow}
\usepackage{times}
\usepackage[dvipsnames]{xcolor}

\newcolumntype{x}[1]{>{\centering\let\newline\\\arraybackslash\hspace{0pt}}p{#1}}

\newcommand{\shape}{\mbox{\boldmath $\beta$}}
\newcommand{\pose}{\mbox{\boldmath $\theta$}}
\newcommand{\vidt}{\mbox{\boldmath $\Theta$}}

\newcommand\blfootnote[1]{%
  \begingroup
  \renewcommand\thefootnote{}\footnote{#1}%
  \addtocounter{footnote}{-1}%
  \endgroup
}
\usepackage[pagebackref=true,breaklinks=true,letterpaper=true,colorlinks,bookmarks=false]{hyperref}

\cvprfinalcopy 
\ifcvprfinal\fi
\begin{document}

\title{Learning 3D Human Dynamics from Video}

\author{Angjoo Kanazawa$^*$, Jason Y. Zhang$^*$, Panna Felsen$^*$, Jitendra Malik\\
University of California, Berkeley\\
{\tt\small\{kanazawa,zhang.j,panna,malik\}@eecs.berkeley.edu}}
\maketitle

\begin{abstract}
From an image of a person in action, we can easily guess the 3D motion of the
person in the immediate past and future. This is because we have a mental model of 3D
human dynamics that we have acquired from observing visual sequences of humans
in motion. We present a framework that can similarly learn a representation of
3D dynamics of humans from video via a simple but effective temporal encoding of
image features. At test time, from video, the learned temporal representation
give rise to smooth 3D mesh predictions. From a single image, our
model can recover the current 3D mesh as well as its 3D past and future
motion. Our approach is designed so it can learn from videos with 2D pose
annotations in a semi-supervised manner.
Though annotated data is always limited, there are millions of videos uploaded daily on the Internet.
In this work, we harvest this Internet-scale source of unlabeled data
by training our model on unlabeled video with pseudo-ground truth 2D pose obtained from an
off-the-shelf 2D pose detector. Our experiments show that adding more videos
with pseudo-ground truth 2D pose monotonically improves 3D prediction performance. 
 We evaluate our model, Human Mesh and Motion Recovery (HMMR), on the recent
 challenging dataset of 3D Poses in the Wild and obtain state-of-the-art
 performance on the 3D prediction task without any fine-tuning.  The project
 website with video, code, and data can be found at {\footnotesize \url{https://akanazawa.github.io/human_dynamics/}}.  \blfootnote{$^*$ equal contribution}
\end{abstract}

\begin{figure}[t]
  \centering
  \includegraphics[width=\columnwidth]{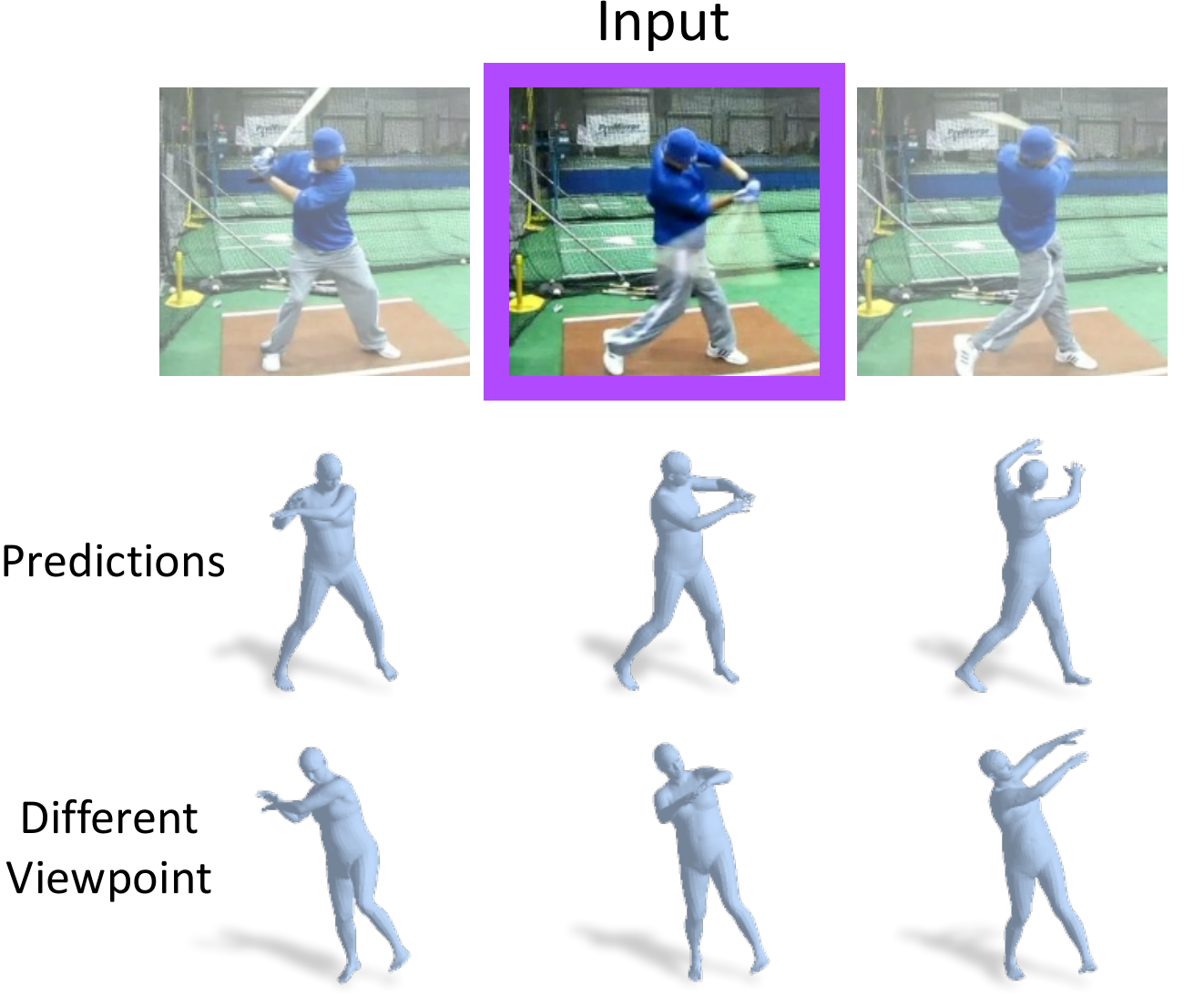}
  \caption{\small{{\bf 3D motion prediction from a single image.} We propose a method that, given a single image of a person, predicts the 3D mesh of the person's body and also hallucinates the future and past motion. Our method can learn from videos with only 2D pose annotations in a semi-supervised manner. 
  Note our training set does not have any ground truth 3D pose sequences of batting motion. Our model also produces smooth 3D predictions from video input.}}
  \vspace{-1.3em}
  \label{fig:teaser}
\end{figure}

\begin{figure*}[t]
  \centering
  \includegraphics[width=0.8\textwidth]{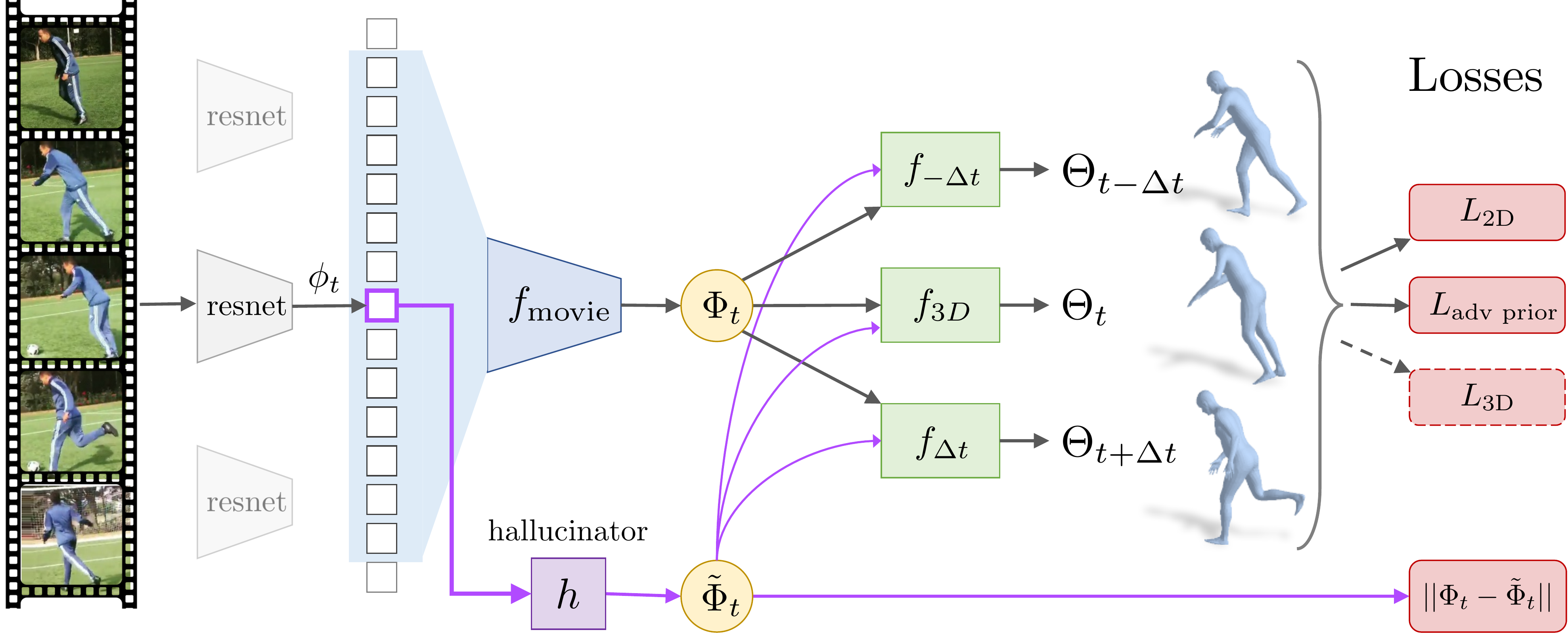}
  \caption{\small{{\bf Overview of the proposed framework: Human Mesh and Motion
        Recovery (HMMR).} Given a temporal sequence of images, we first extract per-image features $\phi_t$. We train a temporal encoder $f_{\text{movie}}$ that learns a representation of 3D human dynamics $\Phi_t$ over the temporal window centered at frame $t$, illustrated in the blue region. From $\Phi_t$, we predict the 3D human pose and shape $\Theta_t$, as well as the change in pose in the nearby $\pm \Delta t$ frames. The primary loss is 2D reprojection error, with an adversarial prior to make sure that the recovered poses are valid. We incorporate 3D losses when 3D annotations are available. We also train a hallucinator $h$ that takes a single image feature $\phi_t$ and learns to hallucinate its temporal representation $\tilde{\Phi}_t$. At test time, the hallucinator can be used to predict dynamics from a single image. 
  }}
  \vspace{-1.3em}
  \label{fig:overview}
\end{figure*}

\vspace{-.5mm}
\section{Introduction}
Consider the image of the baseball player mid-swing in Figure~\ref{fig:teaser}. 
Even though we only see a flat two-dimensional picture, we can infer the
player's 3D pose, as we can easily imagine how his knees bend and arms
extend in space. Furthermore, we can also infer his motion in the surrounding moments as he swings the bat through.
We can do this because we have a mental model of 3D human dynamics that we have
acquired from observing many examples of people in motion.

In this work, we present a computational framework that can similarly learn a model of 3D human dynamics from video.
Given a temporal sequence of images, we first extract per-image features, and then train a simple 1D temporal encoder that learns a representation of 3D human dynamics over a temporal context of image features. We force this representation to capture 3D human dynamics by predicting
not only the current 3D human pose and shape, but also changes in pose in the
nearby past and future frames. We transfer the learned 3D dynamics knowledge to static images by learning a hallucinator that can hallucinate the temporal context representation from a
single image feature. The hallucinator is trained in a self-supervised manner using the
actual output of the temporal encoder. Figure \ref{fig:overview} illustrates the
overview of our training procedure.

At test time, when the input is a video, the temporal encoder can be used to produce
smooth 3D predictions: having a temporal context reduces uncertainty and
jitter in the 3D prediction inherent in single-view approaches. The encoder
provides the benefit of learned smoothing, which reduces the acceleration error by 56\% versus a comparable single-view approach on a recent dataset of 3D humans in the wild. Our approach 
also obtains state-of-the-art 3D error on this dataset without any fine-tuning. When the input
is a single image, the hallucinator can predict the current 3D human mesh as well as 
the change in 3D pose in nearby future and past frames, as illustrated in Figure~\ref{fig:teaser}.

We design our framework so that it can be trained on various types of
supervision. A major challenge in 3D human prediction from a video or an image is
that 3D supervision is limited in quantity and challenging to obtain at a large scale.
Videos with 3D annotations are often captured in a controlled
environment, and models trained on these videos alone do not generalize to the
complexity of the real world. 
When 3D ground truth is not available, our model can be trained with
2D pose annotations via the reprojection loss \cite{3dinterpreter} and an
adversarial prior that constrains the 3D human pose to lie in the manifold of
real human poses \cite{kanazawa18hmr}. 
However, the amount of video labeled with ground truth 2D pose is still limited because ground truth annotations are costly to acquire.

While annotated data is always limited, there are millions of videos uploaded
daily on the Internet. In this work we harvest this potentially unlimited source of unlabeled videos. 
We curate two large-scale video datasets of humans and train on this data using
pseudo-ground truth 2D pose obtained from a state-of-the-art 2D pose detector
\cite{cao2017realtime}. Excitingly, our experiments indicate that adding more videos with pseudo-ground truth 2D monotonically improves the model performance both
in term of 3D pose and 2D reprojection error: 3D pose error reduces by 9\%
and 2D pose accuracy increases by 8\%. Our approach falls in the category of
 omni-supervision \cite{radosavovic2018data}, a subset of semi-supervised
 learning where the learner exploits all data along with Internet-scale unlabeled data. We distill the knowledge of
 an accurate 2D pose detector into our 3D predictors through unlabeled video. While omni-supervision has been
 shown to improve 2D recognition problems, as far as we know, our experiment is
 the first to show that training on pseudo-ground truth 2D pose labels improves 3D
 prediction. 

In summary, we propose a simple but effective temporal encoder that learns to
capture 3D human dynamics. The learned representation allows smooth 3D mesh
predictions from video in a feed-forward manner. The learned representation can
be transferred to a static image, where from a single image, we can predict the
current 3D mesh as well as
the change in 3D pose in nearby frames.
We further show that our model can leverage an Internet-scale source of unlabeled
videos using pseudo-ground truth 2D pose.
    
\section{Related Work}
\vspace{.1em}
\noindent\textbf{3D pose and shape from a single image.}
Estimating 3D body pose and shape from a single image is a fundamentally
ambiguous task that most methods deal by using some model of human bodies and priors. 
Seminal works in this area
\cite{grauman2003inferring,sigal2008combined,agarwal2006recovering} rely on
silhouette features or manual interaction from users \cite{sigal2008combined,guan2009estimating,zhou2010parametric}
to fit the parameters of a statistical body model.
A fully automatic method was proposed by Bogo \etal \cite{SMPLify},
which fits the parametric SMPL \cite{SMPL} model to 2D joint locations detected by
an off-the-shelf 2D pose detector \cite{Leonid2016DeepCut} with strong
priors. Lassner~\etal \cite{UP} extend the approach to fitting predicted silhouettes. \cite{Zanfir} explore the multi-person setting.
Very recently, multiple approaches integrate the SMPL body model
within a deep learning framework
\cite{tung2017self,Tan,pavlakos2018humanshape,kanazawa18hmr,omran2018nbf}, where
models are trained to directly infer the SMPL parameters. These methods vary in the cues they
use to infer the 3D pose and shape: RGB image \cite{Tan,kanazawa18hmr}, RGB image
and 2D keypoints \cite{tung2017self}, keypoints and silhouettes
\cite{pavlakos2018humanshape}, or keypoints and body part segmentations
\cite{omran2018nbf}. Methods that employ silhouettes obtain more accurate
shapes, but require that the person is fully visible and unoccluded in the image. 
Varol \etal explore predicting a voxel representation of
human body \cite{varol18_bodynet}. In this work we go beyond these
approaches by proposing a method that can predict shape and pose from a single image, as well as how the
body changes locally in time. 

\vspace{.5em}
\noindent\textbf{3D pose and shape from video.}
While there are more papers that utilize video, most rely on a multi-view
setup, which requires significant instrumentation. We focus on
videos obtained from a monocular camera.
Most approaches take a two-stage approach:
first obtaining a single-view 3D reconstruction and then post-processing the result
to be smooth via solving a constrained optimization problem \cite{zhou2016sparseness,Wandt,Rehan,rhodin2016general,huang2017towards,VNect_SIGGRAPH2017,2018-TOG-SFV}. Recent methods
obtain accurate shapes and textures of clothing by pre-capturing the actors and making use of silhouettes
\cite{Trumble18,Monoperfcap,Habermann18ReTiCam,alldieck2018video}. While these approaches obtain far more accurate shape,
reliance on the pre-scan and silhouettes restricts these approaches to videos
obtained in an interactive and controlled environments. 
Our approach is
complementary to these two-stage approaches, since all predictions can be
post-processed and refined. There are some recent works that output
smooth 3D pose and shape: \cite{tung2017self} predicts SMPL parameters from two video
frames by using optical flow, silhouettes, and keypoints in a self-supervised
manner. \cite{alldieck2017optical} exploits optical flow to obtain temporally coherent human poses. \cite{totalcapture} fits a body model to a sequence of 3D point clouds and 3D joints obtained from multi-view stereo. Several approaches train LSTM models on various inputs such as image features \cite{lin2017recurrent}, 2D joints \cite{hossain2018exploiting}, or 3D joints \cite{coskun2017long} to obtain temporally coherent 3D joint outputs. 
More recently, TP-Net \cite{dabral2017_tpnet} learns a
fully convolutional network
that smooths the predicted 3D joints. Concurrently to ours, \cite{pavllo:videopose3d:2019} use a fully convolutional network to predict 3D joints from 2D joint sequences. We directly predict the 3D mesh outputs from 2D image sequences and can train with images without any ground truth 3D annotation. 
Furthermore, our temporal encoder
predicts the 3D pose changes in nearby frames in addition to the current 3D pose.
Our experiments indicate that the prediction losses help the encoder to pay more attention to the dynamics information available in the temporal window.

\vspace{.1em}
\noindent\textbf{Learning motion dynamics.}
There are many methods that predict 2D future outputs from video using pixels
\cite{finn2016unsupervised,denton2017unsupervised}, flow \cite{walker2016uncertain}, or
2D pose \cite{pos_iccv2017}. Other methods predict 3D future from 3D inputs \cite{fragkiadaki2015recurrent,jain2016structural,butepage2017deep,li2018auto,villegas2018neural}.
In contrast, our work predicts future and past 3D pose from 2D inputs. 
There are several approaches that predict future from a single image
\cite{walker2015dense,xue2016visual,chao2017forecasting,Prediction-ECCV-2018,gao2018im2flow},
but all approaches predict future in 2D domains, while in this work we propose a
framework that predicts 3D motions. Closest to our work is that of Chao \etal
\cite{chao2017forecasting}, who forecast 2D pose and then estimate the 3D pose from the predicted 2D pose. In this work, we predict dynamics directly in the 3D space and learn the 3D dynamics from video.

\setlength{\abovedisplayskip}{3pt}
\setlength{\belowdisplayskip}{3pt}

\section{Approach}
Our goal is to learn a representation of 3D human dynamics from video, from
which we can 1) obtain smooth 3D prediction and 2) hallucinate 3D motion from static
images.  In particular, we develop a framework that can learn 3D human dynamics from unlabeled, everyday videos of people on the Internet. We first define the
problem and discuss different tiers of data sources our approach can learn from. We then present our framework that
learns to encode 3D human motion dynamics from videos. Finally, we discuss how to transfer this knowledge to
static images such that one can hallucinate short-term human
dynamics from a static image. 
Figure \ref{fig:overview} illustrates the framework.

\subsection{Problem Setup}
Our input is a video $V = \{I_t\}_{t=1}^T$ of length $T$, where each frame is
a bounding-box crop centered around a detected person. We encode the $t$th image frame $I_t$ with a visual feature $\phi_t$,
obtained from a pretrained feature extractor. We
train a function $f_{\text{movie}}$ 
that learns a representation $\Phi_t$ that encodes the 3D dynamics of a human
body given a temporal context of image features centered at frame $t$. Intuitively,
$\Phi_t$ is the representation of a ``movie strip'' of 3D human body in motion at frame $t$. We also learn a hallucinator $h: \phi_t
\mapsto \Phi_t$, whose goal is to hallucinate the movie strip representation
from a static image feature $\phi_t$. 

We ensure that the movie strip representation $\Phi_t$ captures the 3D human body
dynamics by predicting the 3D mesh of a human body from $\Phi_t$ at different
time steps. The 3D mesh of a human body in an image is represented by 85
parameters, denoted by $\Theta = \{\shape, \pose, \Pi\}$,
which consists of shape, pose, and camera parameters. We use the SMPL body 
model \cite{SMPL}, which is a function $\mathcal{M}(\shape, \pose)  \in
\mathbb{R}^{N\times 3}$ that outputs the  $N=6890$ vertices of a triangular mesh
given the shape $\shape$ and pose $\pose$. Shape parameters $\shape \in \mathbb{R}^{10}$ define the linear coefficients of a
low-dimensional statistical shape model, and pose parameters $\pose \in
\mathbb{R}^{72}$ define the global rotation of the body and the 3D relative
rotations of the kinematic skeleton of 23 joints in axis-angle
representation. Please see \cite{SMPL} for more details. The mesh vertices
define 3D locations of $k$ joints $X \in \mathbb{R}^{k \times 3}=W\mathcal{M}(\shape, \pose)$ via
a pre-trained linear regressor $W \in \mathbb{R}^{k \times N}$.
We also solve for the weak-perspective camera $\Pi = [s, t_x,
t_y]$ that projects the body into the image plane. We denote $x = \Pi(X(\shape, \pose))$ as the
projection of the 3D joints.

While this is a well-formed supervised learning task if the ground truth values were available for every
video, such 3D supervision is costly to obtain and not available in
general. Acquiring 3D supervision requires extensive
instrumentation such as a motion capture (MoCap) rig, and these videos captured in a
controlled environment do not reflect the complexity of the real world. While more practical solutions are being
introduced \cite{vonMarcard2018}, 3D supervision is not available for millions of videos
that are being uploaded daily on the Internet. 
In this work, we wish to harness this potentially infinite data source of unlabeled video and
propose a framework that can learn 3D motion from pseudo-ground truth 2D pose
predictions obtained from an off-the-shelf 2D pose detector. Our approach can learn from
three tiers of data sources at once: First, we use the MoCap datasets $\{(V_i, \vidt_i, x_i) \}$
with full 3D supervision $\vidt_i$ for each video along with ground truth 2D
pose annotations for $k$ joints $x_i = \{x_t \in \mathbb{R}^{k \times 2}\}_{t=1}^T$ in each frame. Second, we use datasets of videos in the wild obtained from a monocular camera with human-annotated 2D pose: $\{(V_i, x_i) \}$. 
Third, we also experiment with videos with \emph{pseudo}-ground truth 2D pose: $\{(V_i, \tilde{x}_i) \}$.
See Table \ref{tab:data} for the list of datasets and their details.

\subsection{Learning 3D Human Dynamics from Video}
A dynamics model of a 3D human body captures how the body changes in 3D over
a small change in time. 
Therefore, we formulate this problem as learning a temporal representation that
can simultaneously predict the current 3D body and pose changes
in a short time period. To do this, we learn a temporal encoder $f_{\text{movie}}$ and a 3D regressor $f_{\text{3D}}$ that predict the 3D human mesh
representation at the current frame, as well as delta 3D regressors
$f_{\Delta t}$ that predict how the 3D pose changes in $\pm \Delta t$ time
steps.

\paragraph{Temporal Encoder} Our temporal encoder consists of several layers
of a 1D fully convolutional network $f_{\text{movie}}$ that encodes a temporal
window of image features centered at $t$ into a representation $\Phi_t$ that
encapsulates the 3D dynamics. We use a fully convolutional model for its simplicity. Recent literature also suggests that feed-forward convolutional models empirically out-perform recurrent models
while being parallelizable and easier to train with more stable gradients
\cite{bai2018empirical,miller2018recurrent}. Our temporal convolution network has a ResNet \cite{he2016resnet} based
architecture similar to \cite{bai2018empirical,AfourasCZ18a}.

The output of the temporal convolution network is sent to a 3D regressor
$f_{\text{3D}}:\Phi_t \mapsto \vidt_t$ that predicts the 3D human
mesh representation at frame $t$. We use the same iterative 3D regressor architecture proposed in
\cite{kanazawa18hmr}. Simply having a temporal context reduces ambiguity
in 3D pose, shape, and viewpoint, resulting in a temporally smooth 3D mesh
reconstruction. In order to train these modules from 2D pose annotations, we employ the
reprojection loss \cite{3dinterpreter} and the adversarial prior
proposed in \cite{kanazawa18hmr} to constrain the output pose to lie in
the space of possible human poses. The 3D losses are also used when 3D ground truth is available. Specifically, the loss for the current frame consists of the reprojection loss on visible keypoints 
$
L_{\text{2D}} = ||v_t({x}_t - \hat{x}_t) ||_2^2,
$
where $v_t \in \mathbb{R}^{k \times 2}$ is the visibility indicator over each keypoint, the 3D loss if available,
$
L_{\text{3D}} =  ||\Theta_t - \hat{\Theta}_t||_2^2,
$
and the factorized adversarial prior of \cite{kanazawa18hmr}, which trains a 
discriminator $D_k$ for each joint rotation of the body model $
L_{\text{adv prior}} = \sum_k (D_k(\vidt) - 1)^2.
$

In this work, we regularize the shape predictions using a shape prior $L_\text{$\beta$ prior}$ \cite{SMPLify}.
Together the loss for frame $t$ consists of $L_{t} =
L_{\text{2D}} + L_{\text{3D}} + L_{\text{adv prior}} + L_\text{$\beta$ prior}.$ Furthermore, each sequence is of the same person, so while the pose and
camera may change every frame, the shape remains constant. We express this
constraint as a constant shape loss over each sequence:
\begin{equation}
  \label{eq:shape}
L_{\text{const shape}} = \sum_{t=1}^{T - 1}|| \beta_t - \beta_{t+1} ||.
\end{equation}

\paragraph{Predicting Dynamics}
We enforce that the learned temporal representation captures the 3D human
dynamics by predicting the 3D pose changes in a local time step $\pm\Delta
t$. Since we are training with videos, we readily
have the 2D and/or 3D targets at nearby frames of $t$ to train the dynamics
predictors. Learning to predict 3D changes encourages
the network to pay more attention to the temporal cues, and our experiments show
that adding this auxiliary loss improves the 3D prediction results. Specifically, given a movie strip
representation of the temporal context at frame $\Phi_t$, our goal is to learn a dynamics predictor $f_{\Delta
  t}$ that predicts the change in 3D parameters of the human body at time $t \pm \Delta t$.

In predicting dynamics, we only estimate the change in
3D pose parameters $\theta$, as the shape should remain constant and the
weak-perspective camera accounts for where
the human is in the detected bounding box. In particular, to improve the robustness of the current pose estimation
during training, we augment the image frames with random jitters in scale and translation which
emulates the noise in real human detectors. However, such noise should not be modeled by the dynamics predictor.

For this task, we propose a dynamics predictor $f_{\Delta t}$ that outputs the 72D
change in 3D pose $\Delta \theta$. $f_{\Delta t}$ is a function that maps
$\Phi_t$ and the predicted current pose $\theta_t$
 to the predicted change in pose $\Delta \theta$ for a specific time step
$\Delta t$. 
The delta predictors are trained such that the predicted pose in the
new timestep $\theta_{t + \Delta t} = \theta_t + \Delta \theta$ minimizes the
reprojection, 3D, and the adversarial prior losses at time frame $t + \Delta t$. We use the shape predicted in the current time $t$ to obtain the mesh for $t \pm \Delta t$ frames. To compute the reprojection loss without predicted
camera, we solve for the optimal scale $s$ and
translation $\vec{t}$ that aligns the orthographically projected 3D joints $x_{\text{orth}} = X[:, :2]$
with the visible ground truth 2D joints $x_{gt}$: $\min_{s, \vec{t}}  || (sx_{\text{orth}} + \vec{t}) - x_{gt} ||_2.$
A closed form solution exists for this problem, and we use the optimal camera
$\Pi^* = [s^*, \vec{t}^*]$ to compute the reprojection error on poses predicted
at times $t \pm \Delta t$. 
Our formulation factors away axes of variation, such as shape and
camera, so that the delta predictor focuses on learning the temporal evolution of 3D pose. In summary, the overall objective for the temporal encoder is

\begin{equation}
  L_{\text{temporal}} = \sum_{t}L_{t} + \sum_{\Delta t} L_{t + \Delta t} + L_{\text{const shape}}.\label{eq:movie}
\end{equation}
In this work we experiment with two $\Delta t$ at $\{-5, 5\}$ frames, which
amounts to $\pm 0.2$ seconds for a 25 fps video.

\subsection{Hallucinating Motion from Static Images}
Given the framework for learning a representation for 3D human dynamics,  we now
describe how to transfer this knowledge to static images. The idea is to learn
a hallucinator $h: \phi_t \mapsto \tilde{\Phi}_t$ that maps a single-frame representation
$\phi_t$ to its ``movie strip'' representation $\tilde{\Phi}_t$. One advantage
of working with videos is that during training, the target representation
$\Phi_t$ is readily available for every frame $t$ from the temporal encoder. Thus, the hallucinator can be
trained in a weakly-supervised manner, minimizing the
difference between the hallucinated movie strip and the actual movie strip
obtained from $f_{\text{movie}}$:
\begin{equation}
  \label{eq:hal}
   L_{\text{hal}} = || \Phi_t -  \tilde{\Phi}_t ||_2 .  
\end{equation}
Furthermore, we pass the hallucinated movie strip to the $f_{\text{3D}}$ regressor to
minimize the single-view loss as well as the delta predictors $f_{\Delta t}$.
This ensures that the hallucinated features are not only similar to the actual movie
strip but can also predict dynamics.
All predictor weights are shared among the actual and hallucinated representations.

In summary we jointly train the temporal encoder, hallucinator, and the 
delta 3D predictors together with overall objective:
\begin{equation}
  \label{eq:all}
L = L_{\text{temporal}} + L_{\text{hal}} + L_{t}(\tilde{\Phi}_t) + \sum_{\Delta
  t} L_{t + \Delta t}(\tilde{\Phi}_t).
\end{equation}
See Figure~\ref{fig:overview}
for the overview of our framework.

\section{Learning from Unlabeled Video} \label{sec:unlabeled}
Although our approach can be trained on 2D pose annotations, annotated data is
always limited -- the annotation effort for labeling keypoints in videos is
substantial. However, millions of videos are uploaded to the Internet every
day. On YouTube alone, 300 hours of video are uploaded every minute \cite{YouTubeStats}.

Therefore, we curate two Internet-scraped datasets with pseudo-ground truth
2D pose obtained by running OpenPose \cite{cao2017realtime}. An added advantage
of OpenPose is that it detects toe points, which are not labeled in any of the
video datasets with 2D ground truth. 
Our first dataset is VLOG-people, a subset of the VLOG lifestyle dataset \cite{Fouhey18} on which OpenPose fires consistently. 
To get a more diverse range of human dynamics, we collect another dataset, InstaVariety, from Instagram using 84 hashtags such as \emph{\#instruction}, \emph{\#swimming}, and \emph{\#dancing}. 
A large proportion of the videos we collected contain only one or two people moving with much of their bodies visible, so OpenPose produced reasonably good quality 2D annotations. For videos that contain multiple people, we form our pseudo-ground truth by linking the per-frame skeletons from OpenPose using the Hungarian algorithm-based tracker from Detect and Track \cite{girdhar2018detecttrack}.
A clear advantage of unlabeled videos is that they can be easily
collected at a significantly larger scale than videos with human-annotated
2D pose. Altogether, our pseudo-ground truth data has over 28 hours of 2D-annotated footage, compared to the 79 minutes of footage in the human-labeled datasets. See Table \ref{tab:data} for the full dataset comparison.

\begin{table}[t]  
\centering
     \resizebox{\columnwidth}{!}
     {
\begin{tabular}{x{1.9cm} x{1cm} x{1.1cm}  x{0.85cm}  x{0.87cm} x{0.87cm} x{0.87cm}}
\toprule
\multirow{3}{1.5cm}{\centering Dataset\\Name}  &\multirow{3}{1cm}{\centering Total\\Frames} & \multirow{3}{1cm}{\centering Total\\Length\\(min)}& \multirow{3}{0.85cm}{\centering Avg.\\Length\\(sec)} & \multicolumn{3}{c}{Annotation Type} \\
\cmidrule(lr){5-7}
 &  &  & &GT 3D      & GT 2D    & \small{In-the-wild}     \\ \midrule
Human3.6M                    & 581k                                       & 387                    & 48                          & \checkmark   &  \checkmark     &                \\ \cmidrule(lr){1-7}
Penn Action                   & 77k                                        & 51                      & 3                           &        & \checkmark        & \checkmark                \\
NBA (Ours)                           & 43k                                        & 28                     & 3                          &       & \checkmark        & \checkmark                \\ \cmidrule(lr){1-7}
VLOG peop. & \multirow{2}{*}{353k}            & 236  & \multirow{2}{*}{8}& \multirow{2}{*}{}     & \multirow{2}{*}{}     & \multirow{2}{*}{\checkmark }         \\
(Ours) && (4 hr)&&&\\
  InstaVariety  & \multirow{2}{*}{\textbf{2.1M}} &\textbf{ 1459} & \multirow{2}{*}{6}                   & \multirow{2}{*}{}      & \multirow{2}{*}{}       & \multirow{2}{*}{\checkmark }   \\ 
  
  (Ours)&&\textbf{(1 day)}&&&&\\\bottomrule
\end{tabular} 
}
	\vspace{-0.6em}
\caption{{\small {\bf Three tiers of video datasets.} We jointly train on videos with: full ground truth 2D and 3D pose supervision, only ground truth 2D supervision, and pseudo-ground truth 2D supervision. Note the difference in
  scale for pseudo-ground truth datasets.}}
\vspace{-1.5em}
  \label{tab:data}
\end{table}
\section{Experimental Setup}
\noindent\textbf{Architecture:} We use Resnet-50  \cite{he2016resnet} pretrained on
single-view 3D human pose and shape prediction \cite{kanazawa18hmr} as our
feature extractor, where $\phi_i\in \mathbb{R}^{2048}$ is the the average pooled
features of the last layer. Since training on video requires a large amount of
memory, we precompute the image features on each frame similarly to
\cite{AfourasCZ18a}. This allow us to train on 20 frames of video with
mini-batch size of 8 on a single 1080ti GPU. Our temporal encoder consists of 1D
temporal convolutional layers, where each layer is a residual block of two 1D
convolutional layers of kernel width of 3 with group norm. We use three of these
layers, producing an effective receptive field size of 13 frames. The final output
of the temporal encoder has the same feature dimension as $\phi$. Our hallucinator
contains two fully-connected layers of size 2048 with skip connection. Please see
the supplementary material for more details.

\vspace{.5em}
\noindent\textbf{Datasets:} Human3.6M \cite{Human36m:2014} is the only dataset with ground truth 3D annotations that we train on. It consists of motion capture sequences of actors performing tasks in a controlled lab environment. We follow the standard protocol \cite{kanazawa18hmr} and train on  4 subjects (S1, S6, S7, S8) and test on 2 subjects (S9, S11) with 1 subject (S5) as the validation set. 

For in-the-wild video datasets with 2D ground truth pose annotations, we use
the Penn Action \cite{Zhang2013Penn} dataset and our own NBA dataset. Penn Action
consists of 15 sports actions, with 1257 training videos and 1068 test. We set aside 10\%
of the test set as validation. The NBA dataset contains videos of basketball players
attempting 3-point shots in 16 basketball games. Each sequence contains one set of 2D annotations
for a single player. We split the dataset into 562 training videos, 64 validation, and 151 test.
Finally, we also experiment with the new pseudo-ground truth 2D datasets (Section \ref{sec:unlabeled}).
See Table \ref{tab:data} for the summary of each dataset. Unless otherwise indicated, all models are trained with Human3.6M, Penn Action, and NBA.

We evaluate our approach on the recent 3D Poses in the Wild dataset (3DPW)
\cite{vonMarcard2018}, which contains 61 sequences (25 train, 25 test, 12 val)
of indoor and outdoor activities. Portable IMUs provide ground truth 3D
annotations on challenging in-the-wild videos. To remain comparable to existing methods,
we do not train on 3DPW and only used it as a test set.
For evaluations on all datasets, we skip frames that have fewer than 6 visible keypoints.

As our goal is not human detection, we assume a temporal tube of human detections is available. We use ground truth 2D bounding boxes if available, and otherwise use the output of OpenPose to obtain a temporally smooth tube of human detections. All images are scaled to 224x224 where the humans are roughly scaled to be 150px in height. 

\begin{figure*}[t]
  \centering
  \includegraphics[width=\textwidth]{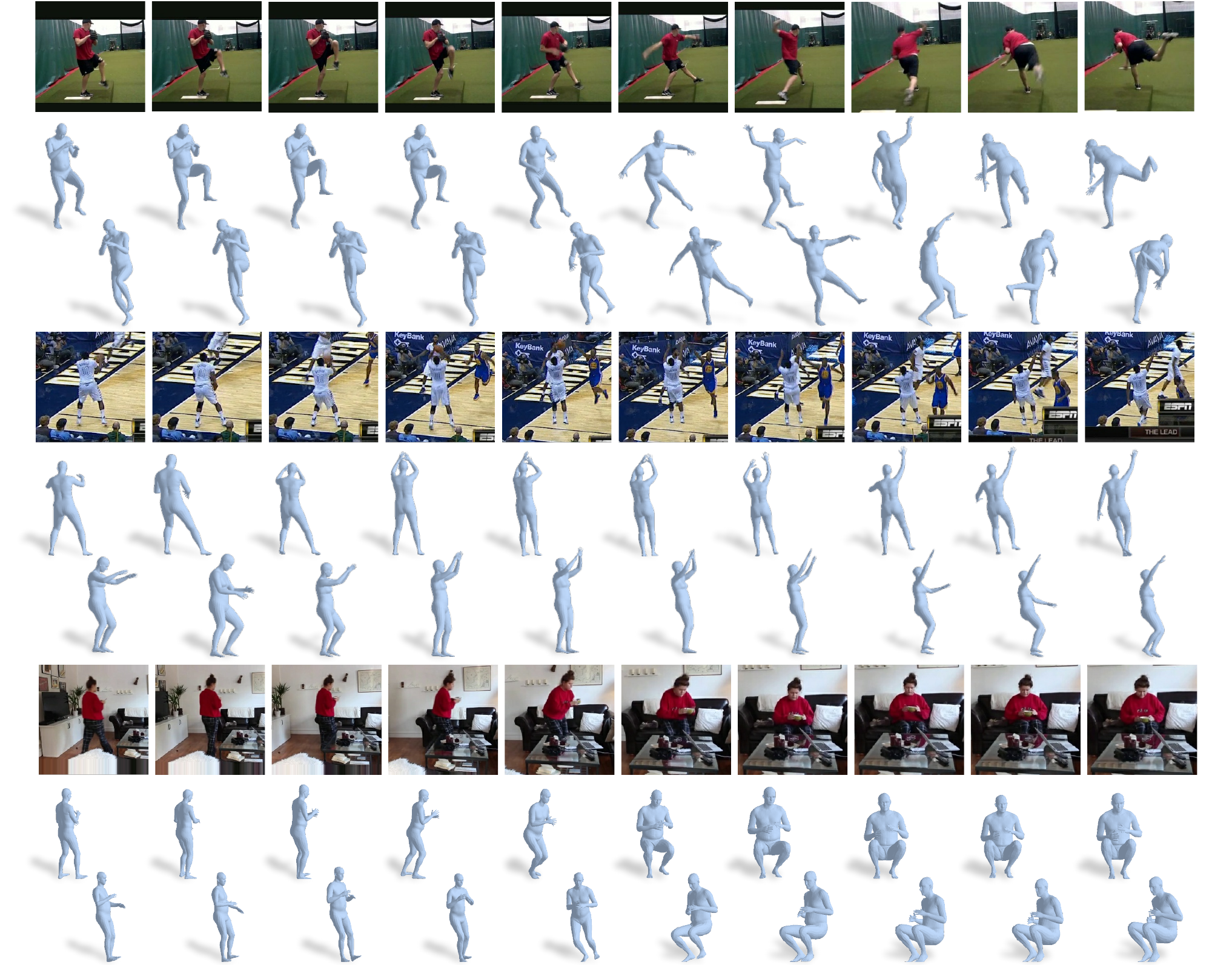}
  \caption{{\bf Qualitative results of our approach on sequences from Penn Action, NBA, and VLOG.} For each sequence, the top row shows the cropped input images, the middle row shows the predicted mesh, and the bottom row shows a different angle of the predicted mesh. Our method produces smooth, temporally consistent predictions.} 
  \vspace{-3mm}
  \label{fig:bigfig}
\end{figure*}
\begin{figure*}[h!]
  \centering
  \includegraphics[width=\textwidth]{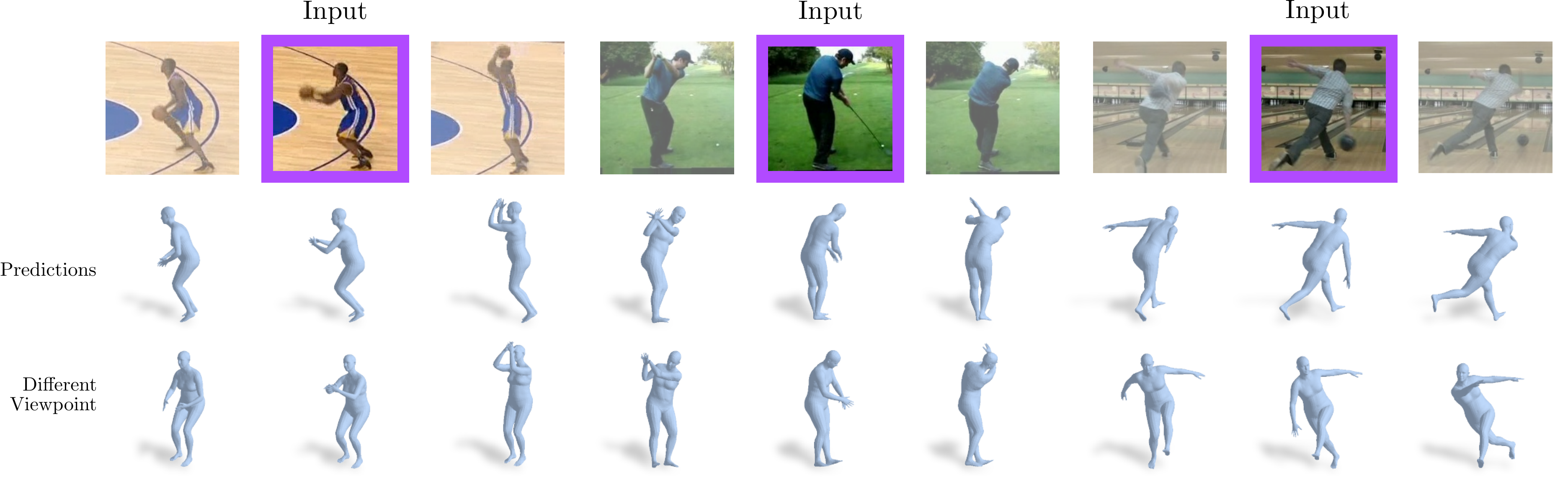}
  \caption{{\bf Predicting 3D dynamics.} In the top row, the boxed image is the single-frame input to the hallucinator while the left and right images are the ground truth past and future respectively. The second and third rows show two views of the predicted meshes for the past, present, and future given the input image.}
  \vspace{-2mm}
  \label{fig:bigdynamics}
\end{figure*}

\section{Experiments}
We first evaluate the efficacy of the learned temporal representation and
compare the model to local approaches that only use a single image. We also
compare our approaches to state-of-the-art 3D pose methods on 3DPW.  We then
evaluate the effectiveness of training on pseudo-ground truth 2D poses. Finally,
we quantitatively evaluate the dynamics prediction from a static image on
Human3.6M. We show qualitative results on video prediction in
Figure~\ref{fig:bigfig} and static image dynamics prediction in
Figure~\ref{fig:teaser} and \ref{fig:bigdynamics}. Please see the supplementary 
for more ablations, metrics, and discussion of failure modes. In addition, a video with more of our results is available at
\url{https://youtu.be/9fNKSZdsAG8}.

\subsection{Local vs Temporal Context}

\begin{table*}[h]
	\centering
	\resizebox{0.9\textwidth}{!}{%
	\begin{tabular}{ccccccccc}
		\toprule
		& \multicolumn{4}{c}{3DPW}                                        & \multicolumn{2}{c}{NBA}                & \multicolumn{2}{c}{Penn Action}        \\
		\cmidrule(lr){2-5} \cmidrule(lr){6-7} \cmidrule(lr){8-9}
		& PCK $\uparrow$ & MPJPE $\downarrow$   & PA-MPJPE $\downarrow$     & Accel Error $\downarrow$   & PCK $\uparrow$             & Accel          & PCK $\uparrow$                   & Accel         \\
		\midrule
		Single-view retrained \cite{kanazawa18hmr}& 84.1 & 130.0 & \textbf{76.7} & 37.4 & 55.9    & 163.6    & 73.2             & 79.9         \\
		Context. no dynamics & 82.6   & 139.2 & 78.4 & \textbf{15.2} & 64.2  & 46.6     & 71.2      & 29.3         \\
		Contextual       & \textbf{86.4} & \textbf{127.1} & 80.1   & 16.4  & \textbf{\textbf{68.4}} & 44.1 & \textbf{77.9} & 29.7 \\ \bottomrule
	\end{tabular}}
	\vspace{-0.6em}
	\caption{\small{{\bf Local vs temporal context.} Our temporal encoder produces
            smoother predictions, significantly lowering the acceleration
            error. We also find that training for dynamic prediction
            considerably improves 2D keypoint estimation.}}
        \vspace{-1.3em}
	\label{tab:mainlocalcontext}
\end{table*}

We first evaluate the proposed temporal encoder by comparing with a single-view approach that only sees a local window of one frame. 
As the baseline for the local window, we use a model similar to
\cite{kanazawa18hmr}, re-trained on the same training data for a fair
comparison. We also run an ablation by training our model with our temporal
encoder but without the dynamics predictions $f_{\Delta t}$.

In order to measure smooth predictions, we propose an \emph{acceleration error},
which measures the average difference between ground truth 3D acceleration and predicted 3D acceleration of each joint in
$mm/s^2$.
This can be computed on 3DPW where ground truth 3D joints are available. On 2D datasets, we simply report the acceleration in $mm/s^2$. 

We also report other standard metrics. For 3DPW, we report the mean per joint
position error (MPJPE) and the MPJPE after Procrustes Alignment (PA-MPJPE). Both
are measured in millimeters. On datasets with only 2D ground truth, we report
accuracy in 2D pose via percentage of correct keypoints \cite{yang2013pck} with $\alpha=0.05$. 

We report the results on three datasets in Table \ref{tab:mainlocalcontext}. Overall, we find that our method produces modest gains in 3D pose estimation,
large gains in 2D, and a very significant improvement in acceleration error. The
temporal context helps to resolve ambiguities, producing smoother, temporally
consistent results. Our ablation study shows that access to temporal context
alone is not enough; using the auxiliary dynamics loss is important to force the
network to learn the \textit{dynamics} of the human. 

\vspace{-.7em}
\paragraph{Comparison to state-of-the-art approaches.}
In Table \ref{tab:main3d}, we compare our approach to other state-of-the-art methods. None of the approaches train on 3DPW. Note that Martinez \etal \cite{martinez_2017_3dbaseline} performs well on the Human3.6M benchmark but achieves the worst performance on 3DPW, showing that methods trained exclusively on Human3.6M do not generalize to in-the-wild images.
We also compare our approach to TP-Net, a recently-proposed semi-supervised approach that is trained on Human3.6M and MPII 2D pose in-the-wild dataset \cite{andriluka14cvpr}. TP-Net also learns a temporal smoothing network supervised on Human3.6M. While this approach is highly competitive on Human3.6M, our approach significantly out-performs TP-Net on in-the-wild video. We only compare feed-forward approaches and not methods that smooth the 3D predictions via post-optimization. Such post-processing methods are complementary to feed-forward approaches and would benefit any of the approaches.

 \begin{table}[t]
	\resizebox{\columnwidth}{!}{%
		
		\begin{tabular}{cccc}
			\toprule
			& \multicolumn{2}{c}{3DPW}                       & \multicolumn{1}{c}{H36M}     \\
			\cmidrule(lr){2-3} \cmidrule(l){4-4}
			& MPJPE $\downarrow$  & \multicolumn{1}{c}{PA-MPJPE $\downarrow$} & \multicolumn{1}{c}{PA-MPJPE $\downarrow$} \\
			\midrule
			Martinez \etal\cite{martinez_2017_3dbaseline} & -               & 157.0                        & 47.7                        \\
			SMPLify \cite{SMPLify}                   & 199.2          & 106.1                        & 82.3                        \\
			TP-Net \cite{Dabral:ECCV:2018}           & 163.7          & 92.3                         & \textbf{36.3}               \\
			Ours                                     & \textbf{127.1} & \textbf{80.1}                & 58.1                        \\
			\cmidrule(lr){1-4}
			Ours + InstaVariety                     & \textbf{116.5} & \textbf{72.6}                & 56.9                        \\ \bottomrule
		\end{tabular}
	}
	\vspace{-0.6em}
	\caption{{\small{\bf Comparison to state-of-the-art 3D pose reconstruction
              approaches.} Our approach achieves state-of-the-art performance on
            3DPW. Good performance on Human3.6M does not always translate to
            good 3D pose prediction on in-the-wild videos.}}
	\label{tab:main3d}
	\vspace{-0.5em}
\end{table}

\subsection{Training on pseudo-ground truth 2D pose}
Here we report results of models trained on the two Internet-scale datasets we collected
with pseudo-ground truth 2D pose annotations (See Table \ref{tab:pseudoGT}). We find that the adding more data
monotonically improves the model performance both in terms of 3D pose and 2D
pose reprojection error. Using the largest dataset, InstaVariety, 3D pose error
reduces by 9\% and 2D pose accuracy increases by 8\% on 3DPW. We see a small
improvement or no change on 2D datasets. It is encouraging to see that not
just 2D but also 3D pose improves from pseudo-ground truth 2D pose
annotations. 

\begin{table}[t!]
\centering
\resizebox{\columnwidth}{!}{%
\begin{tabular}{p{1.7cm}ccccc}
\toprule
                     & \multicolumn{3}{c}{3DPW}                                         & \multicolumn{1}{c}{NBA} & \multicolumn{1}{c}{Penn} \\
                     \cmidrule(lr){2-4} \cmidrule(lr){5-5} \cmidrule(lr){6-6}
                     & PCK  $\uparrow$          & MPJPE $\downarrow$          & \multicolumn{1}{c}{PA-MPJPE $\downarrow$} & PCK $\uparrow$                    & PCK   $\uparrow$ \\ \midrule
 \centering Ours                 & 86.4          & 127.1          & 80.1                         & \textbf{68.4}                    & 77.9                    \\
    
 \centering Ours + VLOG          &\multirow{2}{*}{91.7}& \multirow{2}{*}{126.7}          & \multirow{2}{*}{77.7} & \multirow{2}{*}{68.2}                     & \multirow{2}{*}{78.6}                    \\
 \centering Ours + InstaVariety & \multirow{2}{*}{\textbf{92.9}} & \multirow{2}{*}{\textbf{116.5}} & \multirow{2}{*}{\textbf{72.6}} & \multirow{2}{*}{68.1}   & \multirow{2}{*}{\textbf{78.7}}   \\ \bottomrule
\end{tabular}
}
	\vspace{-0.6em}
\caption{{\small {\bf Learning from unlabeled video via pseudo ground truth 2D pose.} We collected our own 2D pose datasets by running OpenPose on unlabeled video. Training with these pseudo-ground truth datasets induces significant improvements across the board.}}
\vspace{-1.3em}
\label{tab:pseudoGT}
\end{table}

\subsection{Predicting dynamics}
We quantitatively evaluate our static image to 3D dynamics prediction. Since there are no other methods that predict 3D poses from 2D images, we propose two baselines: a constant baseline that outputs the current frame prediction for both past and future, and an Oracle Nearest Neighbors baseline. We evaluate our method on Human3.6M and compare with both baselines in Table \ref{tab:dynamicpred}.

Clearly, predicting dynamics from a static image is a challenging task due to inherent ambiguities in pose and the stochasticity of motion. Our approach works well for ballistic motions in which there is no ambiguity in the direction of the motion. When it's not clear if the person is going up or down our model learns to predict no change.

\begin{table}[]
\centering

\resizebox{0.9\columnwidth}{!}{%
\begin{tabular}{cccc}
	\toprule
	& Past     & Current  & Future   \\ 
	\cmidrule(lr){2-2} \cmidrule(lr){3-3} \cmidrule(lr){4-4}
	& PA-MPJPE $\downarrow$ & PA-MPJPE $\downarrow$ & PA-MPJPE $\downarrow$ \\
	\midrule
	N.N. & 71.6 & \textbf{50.9} & 70.7\\
	Const. & 68.6    & 58.1    & 69.3    \\
	Ours 1  & \textbf{65.0}    & 58.1    & \textbf{65.3}    \\
	Ours 2 & 65.7 & 60.7 & 66.3 \\
	\bottomrule
\end{tabular}
}
	\vspace{-0.6em}
\caption{{\small{\bf Evaluation of dynamic prediction on Human3.6M.} The Nearest Neighbors baseline uses the pose in the training set with the lowest PA-MPJPE with the ground truth current pose to make past and future predictions. The constant baseline uses the current prediction as the future and past predictions. Ours 1 is the prediction model with Eq. \ref{eq:hal}, Ours 2 is that without Eq. \ref{eq:hal}.}}
\vspace{-1.3em}

\label{tab:dynamicpred}
\end{table}

\section{Discussion}
We propose an end-to-end model that learns a model of 3D human dynamics that can
1) obtain smooth 3D prediction from video and 2) hallucinate 3D dynamics on single images
at test time. We train a simple but effective temporal encoder from which the current 3D human
body as well as how the 3D pose changes can be estimated. Our approach can be
trained on videos with 2D pose annotations in a semi-supervised manner, and we
show empirically that our model can improve from training on an
Internet-scale dataset with pseudo-ground truth 2D poses.
While we show promising results, much more remains to
be done in recovering 3D human body from video. Upcoming challenges include
dealing with occlusions and interactions between multiple people. 

\vspace{-1em}

\paragraph{Acknowledgements}
We thank David Fouhey for providing us with the people subset of VLOG, Rishabh Dabral for providing the source code for TP-Net, Timo von Marcard and Gerard Pons-Moll for help with 3DPW, and Heather Lockwood for her help and support. This work was supported in part by Intel/NSF VEC award IIS-1539099 and BAIR sponsors. 

{\small
\bibliographystyle{ieee}
\bibliography{hmmr_references}
}
\clearpage

\section{Appendix}

\subsection{Model architecture}

\paragraph{Temporal Encoder}

\begin{wrapfigure}[15]{r}{0.25\textwidth}
  \vspace{-.2cm}
 \centering
  \includegraphics[height=.65\columnwidth]{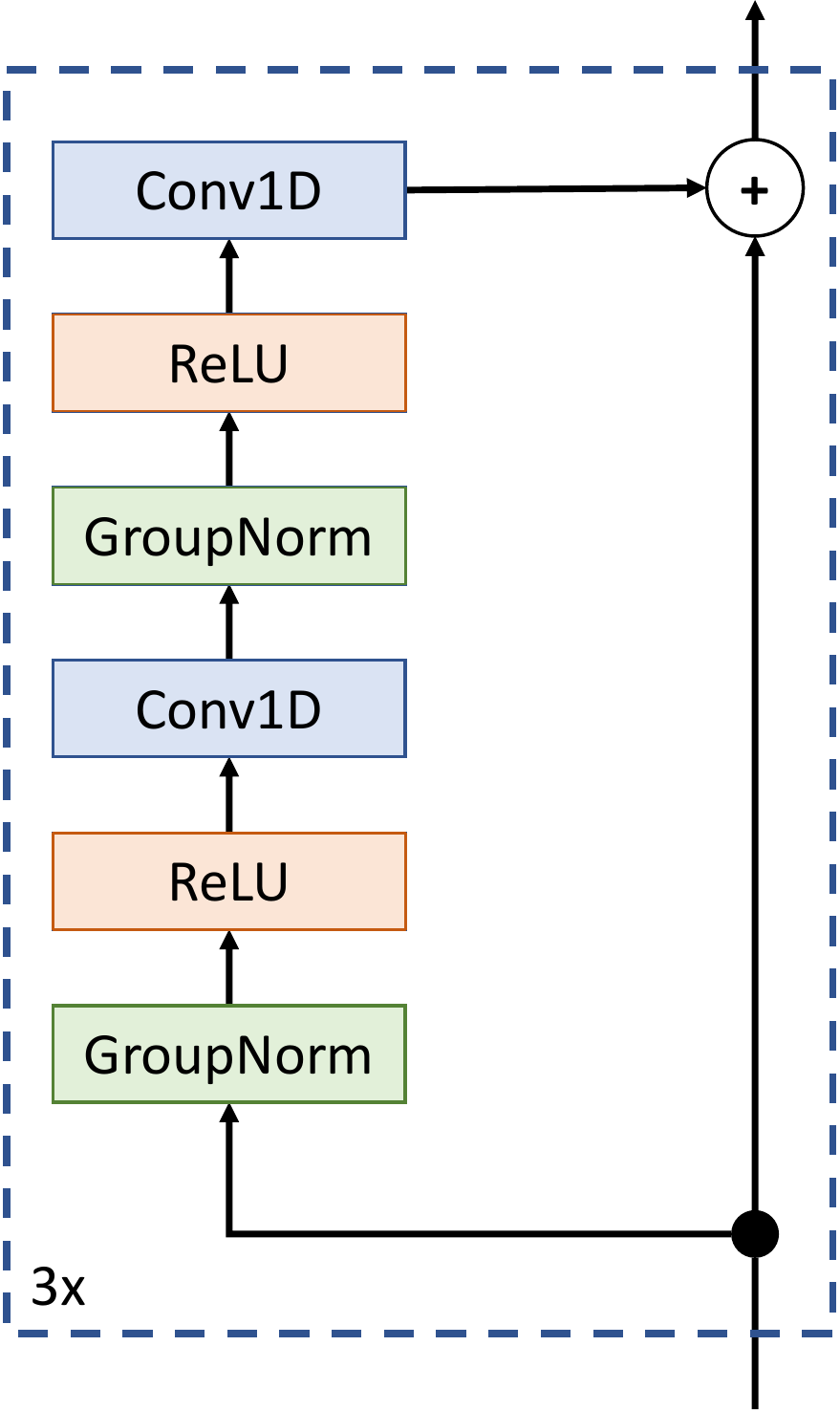}
  \caption{{\small \bf Architecture of $f_{\text{movie}}$.} 
  }
 \label{fig:fmovie}
\end{wrapfigure}
Figure \ref{fig:fmovie} visualizes the architecture of our temporal encoder $f_{\text{movie}}$. Each 1D convolution has temporal kernel size 3 and filter size 2048. For group norm, we use 32 groups, where each group has 64 channels. We repeat the residual block 3 times, which gives us a field of view of 13 frames.  

\paragraph{Hallucinator}
Our hallucinator consists of two fully-connected layers of filter size 2048, whose output gets added to the original $\phi$ as a skip connection. 

\paragraph{3D regressors}
Our $f_{\text{3D}}$ regresses the 85D $\Theta_t$ vector in an iterative error feedback (IEF) loop \cite{IEF,kanazawa18hmr}, where the current estimates are progressively updated by the regressor. Specifically, the regressor takes in the current image feature $\phi_t$ and current parameter estimate $\Theta^{(j)}_t$, and outputs corrections $\Delta \Theta^{(j)}_t$. The current estimate gets updated by this correction $\Theta^{(j+1)}_t  = \Delta \Theta^{(j)}_t + \Theta^{(j)}_t $. This loop is repeated 3 times.
We initialize the $\Theta_t^{(0)}$ to be the mean values $\bar{\Theta}$, which we also update as a part of the learned parameter.

The regressor consists of two fully-connected layers, both with 1024 neurons, with a dropout layer in between, followed by a final layer that outputs the 85D outputs. All weights are shared.

The dynamics predictors $f_{\pm\Delta t}$ has a similar form, except it only outputs the 72-D changes in pose $\theta$, and the initial estimate is set to the prediction of the current frame $t$, \ie 
$\theta_{t+\Delta t}^{(0)} = \theta_t$. Each $f_{\pm \Delta t}$ learns a separate set of weights. 
\subsection{Additional Ablations and Evaluations}

In Table \ref{tab:all_results}, we evaluate our method and comparable methods on 2D/3D pose and 3D shape recovery. We provide another ablation of our approach where the constant shape loss (Eq. 1) is not used (Ours -- Const). In addition, we include full results from our ablation studies.

\paragraph{Shape Evaluation}

To measure shape predictions, we report \textit{Posed Mesh Error} (Mesh Pos), which computes the mean Euclidean distance between the predicted and ground truth 3D meshes. Since this metric is affected by the quality of the pose predictions, we also report \textit{Unposed Mesh Error} (Mesh Unp), which computes the same but with a fixed T-pose to evaluate shape independently of pose accuracy. Both metrics are in units of \textit{mm}.
Note that accurately capturing the shape of the subject is challenging since only 4 ground truth shapes are available in Human3.6M when training. 

\begin{table*}[h]
	\centering
	\resizebox{\textwidth}{!}{%
	\begin{tabular}{ccccccccccc}
		\toprule
		& \multicolumn{6}{c}{3DPW} & \multicolumn{3}{c}{H3.6M}                & \multicolumn{1}{c}{Penn Action}        \\
		\cmidrule(lr){2-7} \cmidrule(lr){8-10} \cmidrule(lr){11-11}
		& PCK $\uparrow$ & MPJPE $\downarrow$   & PA-MPJPE $\downarrow$     & Accel Err $\downarrow$   &  Mesh Pos $\downarrow$ &  Mesh Unp $\downarrow$ &  MPJPE $\downarrow$   & PA-MPJPE $\downarrow$     & Accel Err $\downarrow$   & PCK $\uparrow$  \\
		\midrule
		Martinez \etal\cite{martinez_2017_3dbaseline} &-&-	 & 157.0&-&-		&-		&62.9&	47.7&-&-			\\
		SMPLify \cite{SMPLify} 	& - &199.2&	106.1&	-	&211.2&	61.2&-	&82.3 &- &-\\
		TP-Net \cite{dabral2017_tpnet} &-&	163.7&	92.3&-&-&-& \textbf{52.1} &	\textbf{36.3} & - &-\\
		\cmidrule(lr){1-11}
		Ours & 86.4	&127.1&	80.1&	16.4&	144.4&	25.8&	87.0&	58.1&	9.3 &	77.9\\
		Ours + VLOG & 91.7 &	126.7 &	77.7 &	15.7 &	147.4 &	29.7 &	85.9 &	58.3 &	9.3 &	78.6 \\
		Ours + InstaVariety & \textbf{92.9} & 	\textbf{116.5} & 	\textbf{72.6} & \textbf{	14.3} & 	\textbf{138.6} & 	26.7 & 	83.7 & 	56.9 & 	9.3 & 	\textbf{78.7}\\
		\cmidrule(lr){1-11}
		Single-view retrained \cite{kanazawa18hmr}&84.1	&130.0&	76.7&	37.4	&144.9&	\textbf{24.4}&	94.0&	59.3&	23.9&	73.2\\
		Ours -- Dynamics & 82.6 &	139.2 &	78.4 &	15.2 &	155.2 &	24.8 &	88.6 &	58.3 &	\textbf{9.1} &	71.2\\
		Ours -- Const  &86.5 &	128.3 &	78.2 &	16.6 &	145.9 &	27.5 &	83.5 &	57.8 &	9.3 &	78.1\\
		\bottomrule
	\end{tabular}
	}
	\vspace{-0.6em}
	\caption{\small{{\bf Evaluation of baselines, ablations, and our proposed method on 2D and 3D keypoints and 3D mesh.} We compare with three other feed-forward methods that predict 3D joints. None of the models are trained on 3DPW, all of the models are trained on H3.6M, and only our models are trained on Penn Action (TP-Net also uses MPII 2D dataset). We show that training with pseudo-ground truth 2D annotations significantly improves 2D and 3D predictions on the in-the-wild video dataset 3DPW. Single-view is retrained on our data. Ours -- Dynamics is trained without the past and future regressors $f_{\pm\Delta t}$. Ours -- Const is trained without $L_\text{const shape}$.  }}
        \vspace{-2mm}
	\label{tab:all_results}
\end{table*}

\begin{table*}

 \centering
 \small{
\begin{tabular}{ccccccc}
\toprule
 & \multicolumn{3}{c}{3DPW}  & \multicolumn{2}{c}{H3.6M} & \multicolumn{1}{c}{Penn} \\
 \cmidrule(lr){2-4} \cmidrule(lr){5-6} \cmidrule(lr){7-7}
& PCK $\uparrow$ & MPJPE $\downarrow$ & PA-MPJPE $\downarrow$ & MPJPE $\downarrow$ & PA-MPJPE $\downarrow$ & PCK $\uparrow$ \\ \midrule
Local  & 84.1      & 130.0          & 76.7                         &  94.0     & \textbf{59.3}              & 73.2                    \\    
 \centering Local + Insta. & \textbf{88.6} &    \textbf{126.5} & \textbf{73.0} & \textbf{93.5}   & 59.5 & \textbf{73.2}   \\ 
 Improvement         &5.4\%& 2.7\%                               & 4.8\% & 0.6\% & -0.3\%             & 0\%                    \\
  \cmidrule(lr){1-7}
  \centering Temporal    & 86.4 & 127.1 & 80.1 &  87.0 & 58.1   & 77.9    \\    
 \centering Temp. + Insta. & \textbf{92.9} &   \textbf{116.5} & \textbf{72.6} & \textbf{83.7} & \textbf{56.9}   & \textbf{78.7}   \\ 
 \centering Improvement         &\textbf{7.6\%}& \textbf{8.4\%} & \textbf{9.3\%} &\textbf{3.7\%} & \textbf{2.0\%} & \textbf{1.0\%} \\
\bottomrule  

\end{tabular}
}
	\vspace{-0.6em}
\caption{{\small {\bf Effects of using unlabeled video via pseudo ground truth
      2D pose on local (single-view) vs temporal context models.} While both
    local and temporal context models benefit from training with additional
    pseudo-ground truth datasets in general, the gain is more significant with the model
    with temporal context. A possible explanation is that the temporal context acts as a
    regularizer that allows it to be more robust to outlier poor 2D pseudo-ground labels.}}
\label{tab:insta}
\end{table*}

\subsection{Failure Modes}
While our experiments show promising results, there is still room for improvement.
\vspace{-4mm}
\paragraph{Smoothing}
Overall, our method obtains smooth results, but it can struggle in challenging situations, such as person-to-person occlusions or fast motions. Additionally, extreme or rare poses (\eg stretching, ballet) are difficult to capture. Please refer to our supplementary video for examples.

\begin{figure}[h]
  \centering
  \includegraphics[height=.7\columnwidth]{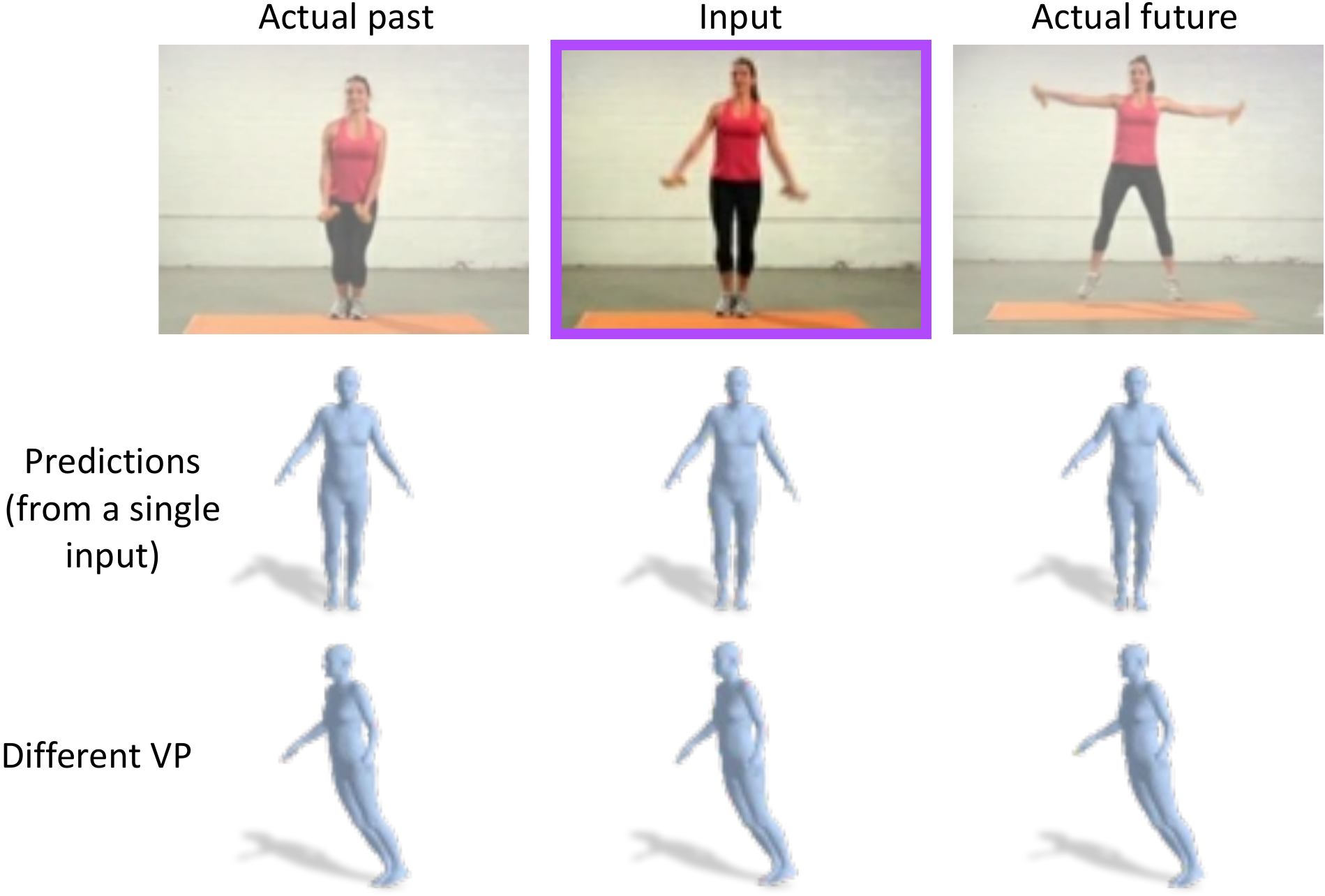}
  \caption{{\bf Ambiguous motion}. Dynamic prediction is difficult from the center image alone, where her arms may reasonably lift or lower in the future.
  }
  \label{fig:dynamic_fail}
\end{figure}

\vspace{-4mm}
\paragraph{Dynamics Prediction}
Clearly, predicting the past and future dynamics from a single image is a challenging problem. Even for us humans, from a single image alone, many motions are ambiguous. Figure \ref{fig:dynamic_fail} visualizes a canonical example of such ambiguity, where it is unclear from the input, center image, if she is about to raise her arms or lower them. In these cases, our model learns to predict constant pose. 

\begin{figure}[h]
  \centering
  \includegraphics[height=.7\columnwidth]{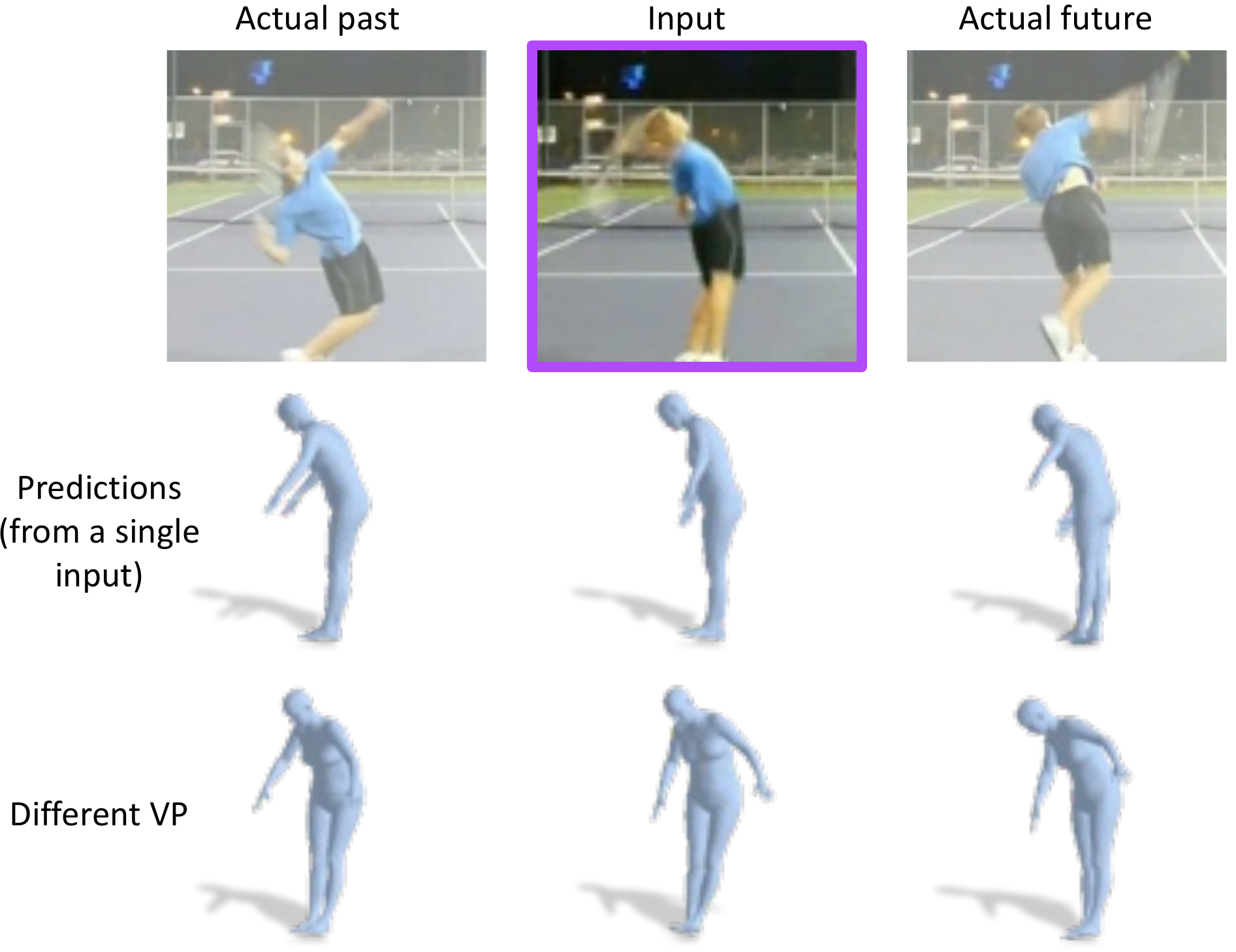}
  \caption{{\bf Ambiguous pose}. The tennis player's pose in the input, center image is difficult to disambiguate between hunched forward verses arched backward due to the motion blur. This makes it challenging for our model to recover accurate dynamics predictions from the single image.
  }
  \vspace{-.5em}
  \label{fig:dynamic_fail2}
\end{figure}

Furthermore, even the pose in a single image can be ambiguous, for example due to motion blur in videos. Figure \ref{fig:dynamic_fail2} illustrates a typical example, where the tennis player's arm has disappeared and therefore the model cannot discern whether the person is facing left or right. When the current frame prediction is poor, the resulting dynamics predictions are also not correct, since the dynamics predictions are initialized from the pose of the current frame.

Note that incorporating temporal context resolves many of these static-image ambiguities. Please see our included supplementary video for examples.

\subsection{Effects of pseudo-ground truth dataset on local vs temporal models}
Here we look further into the effects of adding the large-scale, unlabeled InstaVariety dataset with
pseudo-ground truth 2D annotations. The main paper showed that adding
the large scale InstaVariety dataset with pseudo-ground truth 2D labels improved
the performance significantly.
Here we analyze how much the temporal
context helps in learning from noisy pseudo-ground truth annotations, by training
the local, single-view model also with the extra InstaVariety dataset. The
results are shown in Table \ref{tab:insta}. While both models benefit from
pseudo-ground truth datasets, we find the relative improvement is significantly higher for the
temporal model. This suggests that temporal context allows the model to be more
robust toward bad, outlier pseudo-ground truth 2D labels.

\end{document}